 \documentclass[final,5p,times,twocolumn,authoryear]{elsarticle}

\usepackage{amssymb}
\usepackage{lipsum}
\usepackage{float}

\journal{Automation in Construction}

\begin{document}

\begin{frontmatter}



\title{Developing a Comprehensive Measurement Tool for Assessing the Rate of BIM Adoption in the Construction Industry}


\author[first]{Mohammed Abdulsalam Alsofiani, Ph.D}
\affiliation[first]{organization={Texas A\&M University} 
           }

\begin{abstract}
Building Information Modeling (BIM) is a crucial technology in the construction industry, offering benefits such as enhanced collaboration, real-time decision-making, and significant cost and time savings. Despite its advantages, BIM adoption faces numerous barriers. This study aims to create a reliable tool to assess the Rate of BIM Adoption (RBA), drawing on Attributes of Innovation theory and empirical data from the literature. 
This research integrates theoretical insights with empirical data, providing quantitative items to measure BAR in the construction industry. The quantitative approach helps decision-makers and policymakers to mandate BIM and establish appropriate implementation standards. Its implications are significant for the construction industry, policymakers, and the academic community, offering a systematic approach to assess BIM adoption, identify barriers, and implement targeted strategies. The reliability of this approach is ensured through a solid theoretical foundation, item development, pilot testing, and statistical analysis, making it a valuable resource for improving BIM implementation and fostering innovation in the construction industry.

\end{abstract}




\end{frontmatter}




\section{Introduction}

When it comes to technology adoption, particularly Building Information Modeling (BIM), many variables become critical to consider. However, the perception of BIM adoption needs to be numerically measured to evaluate the rate of adoption. Hence, there is a need to develop a measurement tool for the Rate of BIM Adoption (RBA). However, despite these advantages, the adoption of BIM is often hindered by several barriers. To understand and quantify the rate at which BIM is adopted, it is essential to develop a reliable measurement tool.
This article aims to address the need for a comprehensive measurement tool for the Rate of BIM Adoption (RBA). It draws on the theoretical framework of the Attributes of Innovation by Rogers (2003) and builds on the systematic review conducted by Alsofiani (2024), which examined BIM adoption barriers in infrastructure construction projects. Additionally, Rogers' theory, widely recognized for explaining the adoption of innovations, provides a sociological perspective on the factors that affect how quickly new technologies are embraced. Yet, the attributes of innovation can be categorized and modified based on the nature of the technology intended to be studied \citep{damanpour1991organizational,dibra2015rogers}. Moreover, the article incorporates Rogers' (2003) identified five stages of the innovation-decision process and the strategies a change agent can use to impact each stage.  By integrating these insights, this study provides quantitative instrument items that can measure the degree to which the construction industry fully adopts BIM technologies. This tool can assist decision-makers and policymakers in mandating BIM and establishing suitable implementation and project delivery standards.
By combining these theoretical insights with empirical data on the barriers to BIM adoption, this article seeks to create a tool that can be used by the scientific community to measure the RBA effectively. Furthermore, the attributes of innovation can be tailored to fit the specific characteristics of BIM, ensuring that the measurement tool is both relevant and robust. This research not only contributes to the academic understanding of BIM adoption but also provides practical implications for improving BIM implementation strategies in the construction industry.
\section{Background}
\subsection{Benefits and Barriers of BIM Adoption}
The study "Digitalization in Infrastructure Construction Projects: A PRISMA-Based Review of Benefits and Obstacles" by \cite{alsofiani2024digitalization} explores the advantages and barriers associated with the adoption of Building Information Modeling (BIM) in infrastructure projects over a decade (2013-2023). BIM offers significant benefits throughout the project life cycle, including planning, design, construction, maintenance, and sustainability. These benefits include enhanced collaboration, real-time data-driven decision-making, and substantial cost and time savings.\\
The systematic review identified 74 barriers to BIM adoption from 11 in-depth studies. These barriers were categorized into seven primary impediments: education/training challenges, resistance to change, unclear business value, perceived costs, lack of standards and guidelines, lack of mandates, and lack of initiatives. The study emphasizes the need for a comprehensive measurement tool for the Rate of BIM Adoption (RBA) to quantify BIM adoption effectively and develop strategies to overcome these barriers.\\
The review highlights the critical role of BIM in improving project efficiencies and suggests that understanding the benefits and barriers is essential for stakeholders. Identifying the benefits helps organizations make informed decisions about BIM adoption, leading to increased project efficiency, cost savings, and improved outcomes. Conversely, recognizing barriers enables the development of practical strategies to overcome obstacles and facilitate BIM adoption.\\
The study's methodology involved two key stages. The first stage identified the advantages of implementing BIM in infrastructure projects using data visualization tools like VOSviewer. This tool illustrated co-occurrence patterns and relationships among keywords and authors in the selected articles. The second stage involved a content analysis following the PRISMA protocol, which provided a framework for evidence-based reporting. The qualitative synthesis and categorization of identified barriers aimed to develop a comprehensive understanding of the primary challenges associated with BIM adoption.\\
The study concludes that while BIM adoption offers numerous benefits, several challenges need to be addressed to fully realize its potential. Future research directions include enhancing education and training, promoting standardization, advocating for policies and mandates, and integrating BIM with emerging technologies. The study underscores the importance of addressing these barriers to pave the way for more efficient, sustainable, and cost-effective construction practices. This research not only contributes to academic understanding but also provides practical implications for improving BIM implementation strategies in the construction industry.

\subsection{Attributes of Innovation Theory}
According to \cite{rogers2014diffusion}, the rate of adoption is defined as the relative speed with which an invention is accepted by members of a social system. The popularity of a new idea is often assessed by the number of people who embrace it within a defined period of time. Therefore, the rate of adoption of an invention is a numerical measure of how steep the adoption curve for that innovation will be. Thus, Rogers developed the diffusion of innovation theory, which has since become the most accepted theory for explaining users’ behaviors at which technology adapts to a new environment. \\
According to Diffusion of Innovation theory, the rate of adoption is determined by evaluating the five attributes of innovation: relative advantage, compatibility, complexity, trialability, and observability. However, according to \cite{rogers2014diffusion}, the rate of adoption of an innovation is also influenced by a number of other factors, including: (a) the type of innovation decision made; (b) the nature of the communication channels that are used to disseminate the innovation at various stages in the innovation-decision process; (c) the nature of the social system; and (d) the extent to which change agents are involved in disseminating an innovation. \\
\cite{rogers2014diffusion} outlined five main factors that impact a person’s decision to adopt or reject an innovation as shown in Figure 1, which can be used to impact the rate of adoption: 
\begin{enumerate}
    \item \textbf{Relative advantage}: how much better an invention is compared to its predecessor.
    \item \textbf{Compatibility}: the degree to which an invention must be compatible with the lifestyle of a person for it to be adopted.
    \item \textbf{Complexity}: Individuals are less likely to accept new technologies when they are too complex to use.
    \item \textbf{Trialability}: the level at which innovative concepts may be readily adopted.
    \item \textbf{Observability}: how noticeable an invention is to others (more prominent innovation generates more favorable or unfavorable responses among a population)
\end{enumerate}

\begin{figure} [H]
	\centering 
	\includegraphics[width=0.4\textwidth, angle=-0]{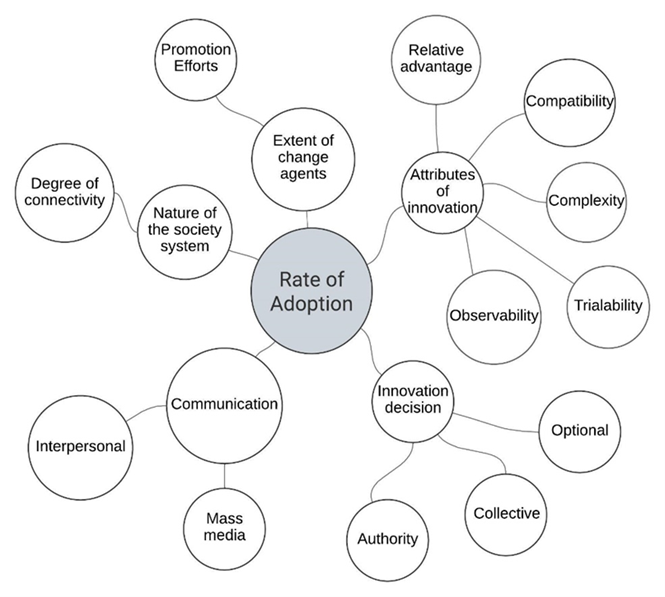}	
	\caption{Variables Determining the Rate of Adoption of Innovation Adapted from Attributes of Innovation Theory} 
	
\end{figure}
\begin{figure*} [!ht]

	\centering 
	\includegraphics[width=.8\textwidth, angle=-0]{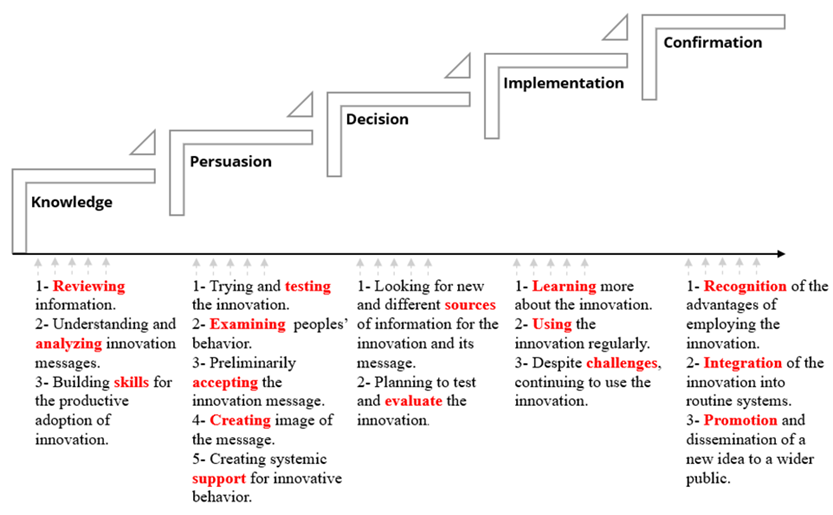}	
	\caption{Strategies for Change Agent to Impact the Adoption Decision} 
	
\end{figure*}
\subsection{Change Agent Strategies }
There are several strategies a change agent can use to impact the adoption decision process as shown in Figure 2. \cite{rogers2014diffusion} identified five stages for the innovation decision process and proposed a set of strategies to help achieve the core objectives of each stage.
The first stage is knowledge. Three strategies can be used to positively impact the development of knowledge: (a) reviewing information, (b) understanding and analyzing innovation messages, and (c) building skills for the productive adoption of innovation. The second stage is persuasion. Persuasion encompasses a set of five strategies: (a) trying and testing the innovation, (b) examining how other people react to the new behavior, (c) preliminarily accepting the innovation message, (d) creating a favorable image of the message and innovation, and (e) creating systemic support for innovative behavior. The third stage involves the decision, which is the most critical stage in the innovation-decision process. At this stage, two strategies are effective for constructing a solid adoption decision: (a) looking for new and different sources of information for the innovation and its message, and (b) planning to test and evaluate the innovation. The fourth stage is implementation. Three strategies can be used to enhance the implementation process: (a) learning more about the innovation, (b) using the innovation regularly, and (c) despite challenges, continuing to use the innovation. The fifth and final stage is confirmation. In this stage, the adopted decision must be reinforced. Three reinforcement strategies are effective at this point: (a) recognition of the advantages of employing the innovation, (b) integration of the innovation into 
routine systems, and (c) promotion and dissemination of a new idea to a wider public.

\section{Methodology}
This study employs a systematic methodology to develop a reliable measurement tool for the Rate of BIM Adoption (RBA) in the construction industry. The methodology integrates Rogers' Attributes of Innovation theory and the findings from a systematic review conducted by \cite{alsofiani2024digitalization}, which examined barriers to BIM adoption in infrastructure construction projects. The following steps outline the process:

\subsection{Theoretical Foundation}
The theoretical basis for the measurement tool is Rogers' Attributes of Innovation theory. This theory identifies five key attributes that influence the adoption of innovations: relative advantage, compatibility, complexity, trialability, and observability. These attributes provide a comprehensive framework for understanding the factors that impact the adoption of BIM in the construction industry.

\subsection{Systematic Literature Review}
Building on the systematic review conducted by \cite{alsofiani2024digitalization}, this study incorporates empirical findings to identify barriers to BIM adoption. The systematic review followed a rigorous and transparent process adhering to the PRISMA (Preferred Reporting Items for Systematic Reviews and Meta-Analyses) guidelines:

\begin{enumerate}
    \item \textbf{Search Strategy}: Relevant databases, including Web of Science and ScienceDirect, were searched using keywords such as "Building Information Modeling," "BIM adoption," and "infrastructure projects." The search was confined to peer-reviewed journal articles published between 2013 and 2023.
    \item \textbf{Inclusion and Exclusion Criteria}: Articles were included if they focused on BIM adoption in infrastructure projects, were published in English, and underwent peer review. Exclusions were made for duplicate records, conference papers, books, and reports.
    \item \textbf{Screening and Selection}: Articles were screened based on titles and abstracts, with relevant studies selected for full-text review.
    \item \textbf{Data Extraction and Synthesis}: Data were extracted on identified barriers and benefits of BIM adoption. These data were synthesized to identify common themes and categories, culminating in the identification of 74 barriers grouped into seven primary impediments.
    \item 
\end{enumerate}

\subsection{Item Development}

Based on Rogers' theoretical framework and the systematic review findings, Likert scale statements were developed to measure key variables influencing BIM adoption. The statements were categorized into the following groups:
\begin{enumerate}
    \item \textbf{Relative Advantage}: Assesses perceived benefits of BIM over traditional methods.
    \item \textbf{Compatibility}: Evaluates how well BIM integrates with existing processes and goals.
    \item \textbf{Complexity}: Measures the perceived ease or difficulty of learning and using BIM technology.
    \item \textbf{Trialability}: Assesses the extent to which BIM can be tested on a small scale before full implementation.
    \item \textbf{Observability}: Focuses on the visibility of the benefits of BIM adoption.
    \item \textbf{Additional Barriers}: Identifies specific obstacles that hinder BIM adoption.
\end{enumerate}
Each category includes multiple statements to capture different aspects of the variable being measured.

\section{Findings}
The development of a comprehensive measurement tool for the Rate of BIM Adoption (RBA) necessitates an understanding of various factors that influence the adoption process. This study categorizes these factors into several key areas based on Rogers' Attributes of Innovation theory and insights from a systematic review conducted by \cite{alsofiani2024digitalization}. As shown in Table 1, the findings (items) are developed to measure the degree of agreement on six item groups:
relative advantage, compatibility, complexity, trialability, observability, and additional barriers.
\begin{table*}[!ht]
\centering
\begin{tabular}{l l} 
 \hline
 \textbf{Category} & \textbf{Likert Scale Statements to Assess the Rate of BIM Adoption in the Construction Industry} \\ 
 \hline
 Relative Advantage & 1. BIM adoption significantly improves project efficiency and productivity. \\
 & 2. Implementing BIM leads to substantial cost savings in construction projects. \\
 & 3. BIM enhances the quality and accuracy of project designs. \\
 & 4. Using BIM technology reduces the number of errors and rework in projects. \\
 \hline
 Compatibility & 5. BIM tools are compatible with our existing project management processes. \\
 & 6. The integration of BIM aligns well with our company's goals and strategies. \\
 & 7. BIM adoption fits seamlessly with our current technological infrastructure. \\
 & 8. BIM is suitable for the types of projects we typically undertake. \\
 \hline
 Complexity & 9. Learning to use BIM software is straightforward and user-friendly. \\
 & 10. The complexity of BIM tools hinders their adoption in our projects. (Reverse-scored) \\
 & 11. Our team finds BIM technology easy to implement and use. \\
 & 12. The training required to use BIM is manageable for our staff. \\
 \hline
 Trialability & 13. We have had sufficient opportunities to test BIM before full implementation. \\
 & 14. Piloting BIM on smaller projects has been beneficial for understanding its advantages. \\
 & 15. Trial runs with BIM have positively influenced our decision to adopt it. \\
 & 16. The ability to experiment with BIM tools has increased our confidence in using them. \\
 \hline
 Observability & 17. The benefits of BIM adoption are clearly visible in our completed projects. \\
 & 18. Successful BIM implementation in other companies is evident and convincing. \\
 & 19. Positive outcomes from BIM adoption are easily observable in our industry. \\
 & 20. The advantages of using BIM are apparent in the projects where it has been applied. \\
 \hline
 Additional Barriers & 21. Lack of standardized guidelines for BIM implementation is a significant barrier. \\
 & 22. Resistance to change within the organization impedes BIM adoption. \\
 & 23. The perceived high cost of adopting BIM limits its usage. \\
 & 24. Insufficient education and training opportunities for BIM are a challenge. \\
 & 25. Lack of government mandates for BIM adoption affects our decision to implement it. \\
 & 26. Unclear business value of BIM prevents us from fully embracing it. \\
 & 27. Limited technical support for BIM adoption hinders its implementation. \\
 \hline
\end{tabular}
\caption{Likert Scale Statements for Measuring Rate of BIM Adoption (RBA)}
\label{Table1}
\end{table*}

\subsection{Relative Advantage}
The relative advantage category assesses the perceived benefits of BIM over traditional methods. This includes improvements in project efficiency and productivity, cost savings, enhanced quality and accuracy of designs, and reduction in errors and rework. Understanding the relative advantage of BIM helps stakeholders to justify its adoption by highlighting tangible benefits that contribute to better project outcomes.
\subsection{Compatibility}
The compatibility category evaluates how well BIM integrates with existing processes, goals, and technological infrastructure of an organization. Statements in this category measure whether BIM fits with current project management practices, aligns with company strategies, and is suitable for the types of projects undertaken. Compatibility is crucial for seamless adoption and minimizing resistance to change.
\subsection{Complexity}
The complexity category measures the perceived ease or difficulty of learning and using BIM technology. This includes the user-friendliness of BIM software, the manageability of required training, and whether the complexity of BIM tools poses a barrier to adoption. Understanding complexity helps identify areas where additional support or training may be needed to facilitate smoother implementation.
\subsection{Trialability}
The trialability category assesses the extent to which BIM can be tested on a small scale before full implementation. This includes opportunities for piloting BIM on smaller projects, the influence of trial runs on adoption decisions, and the impact of experimentation on confidence in using BIM. Trialability is important for building trust and familiarity with the technology, reducing perceived risks.
\subsection{Observability}
The observability category focuses on the visibility of the benefits of BIM adoption. This includes the clarity of BIM's positive outcomes in completed projects, the evident success of BIM in other companies, and the overall visibility of its advantages in the industry. High observability can encourage wider acceptance and adoption by showcasing concrete examples of BIM’s effectiveness.
\subsection{Additional Barriers}
The additional barriers category identifies specific obstacles that hinder BIM adoption. These include lack of standardized guidelines, resistance to change, perceived high costs, insufficient education and training opportunities, lack of government mandates, unclear business value, and limited technical support. Addressing these barriers is essential for facilitating smoother and more widespread adoption of BIM.
\section{Reliability}

Reliability is a critical aspect of the measurement tool developed for assessing the Rate of BIM Adoption (RBA). Ensuring that the tool consistently produces accurate and stable results over time is essential for its effectiveness and credibility. Several steps have been taken to enhance the reliability of this measurement tool:
\subsection{Systematic Review and Theoretical Foundation}
The tool is grounded in a robust theoretical framework, drawing on the Attributes of Innovation by Rogers (2003) and a systematic review of barriers to BIM adoption identified by Alsofiani (2024). This dual foundation ensures that the variables and statements included in the tool are well-supported by existing literature and theory, enhancing their validity and reliability.
\subsection{Comprehensive Item Development}
The Likert scale statements were developed based on a thorough review of the relevant literature and theoretical models. Each statement was carefully crafted to ensure clarity, relevance, and alignment with the identified attributes of innovation and barriers to BIM adoption. This comprehensive approach minimizes ambiguity and enhances the precision of the measurement tool.
\subsection{Pilot Testing and Refinement}
To ensure the reliability of the measurement tool, it should undergo pilot testing with a sample of construction industry professionals. This initial testing phase will help identify any inconsistencies or ambiguities in the statements. Feedback from the pilot testing will be used to refine and adjust the tool, ensuring that it accurately captures the intended variables and produces consistent results across different contexts and respondents.

\section{Implication}

The findings of this study have significant implications for the construction industry, policymakers, and the academic community. By developing a comprehensive measurement tool for the Rate of BIM Adoption (RBA), this research offers several practical and theoretical contributions that can enhance BIM implementation strategies and improve overall project outcomes.
\subsection{For the Construction Industry}
The measurement tool provides a systematic approach for organizations to assess their BIM adoption rate, identify critical barriers, and implement targeted strategies to overcome these challenges. By understanding the relative advantages, compatibility, complexity, trialability, and observability of BIM, construction firms can make informed decisions about investing in and integrating BIM technologies. The tool also highlights the importance of addressing additional barriers such as lack of standardized guidelines, resistance to change, perceived high costs, and insufficient education and training opportunities. Implementing the change agent strategies across the five stages of the innovation-decision process can further facilitate successful BIM adoption and integration into routine practices.
\subsection{For Policymakers}
Policymakers can use the insights from this study to develop and enforce regulations and standards that support BIM adoption. The measurement tool can help in evaluating the effectiveness of existing policies and identifying areas that require additional support or intervention. By mandating BIM use and promoting standardized guidelines, policymakers can drive industry-wide adoption and ensure that the benefits of BIM are realized across all infrastructure projects. Additionally, investing in education and training programs can address skill gaps and prepare the workforce for successful BIM implementation.
\subsection{For the Academic Community}
This research contributes to the theoretical understanding of technology adoption in the construction industry by integrating Rogers' Attributes of Innovation with empirical data on BIM adoption barriers. The development of quantitative instrument items provides a robust framework for future studies to measure BIM adoption rates and evaluate the impact of various factors on the adoption process. Academics can build on this research to explore further the specific challenges and opportunities associated with BIM adoption in different contexts and regions. The study also opens avenues for longitudinal research to track the long-term effects of BIM implementation on project performance and sustainability.
Overall, the implications of this study extend beyond the immediate context of BIM adoption in construction projects. By providing a comprehensive measurement tool and highlighting effective change agent strategies, this research offers valuable insights for improving technology adoption processes, enhancing project efficiencies, and fostering innovation in the construction industry.

\section{Conclusion}
This article addresses the critical need for a comprehensive measurement tool for the Rate of BIM Adoption (RBA) in the construction industry. By drawing on the theoretical framework of the Attributes of Innovation by \cite{rogers2014diffusion} and incorporating insights from the systematic review conducted by \cite{alsofiani2024digitalization}, the study identifies and quantifies the key variables that influence BIM adoption. The integration of Rogers' theory provides a sociological perspective on the factors affecting the adoption process, which can be tailored to the specific characteristics of BIM technology.
The research highlights several barriers to BIM adoption, including lack of standardized guidelines, resistance to change, perceived high costs, insufficient education and training opportunities, lack of government mandates, unclear business value, and limited technical support. Addressing these barriers is essential for facilitating smoother and more widespread adoption of BIM.
Additionally, the article emphasizes the importance of change agent strategies across the five stages of the innovation-decision process: knowledge, persuasion, decision, implementation, and confirmation. By employing targeted strategies at each stage, organizations can enhance their adoption processes and achieve greater success with BIM implementation.

\section*{Acknowledgements}
This research is based on my doctoral work conducted at Texas A\&M University. I would like to thank my advisors, Dr. Stephen Caffey, Dr. Edelmiro Escamilla, Dr. Michael Lewis, and Dr. Kim Dooley, for their guidance and support throughout this project. Additionally, I would like to express my deep appreciation to my country, Saudi Arabia, for funding and supporting me during this journey.

\appendix








\begin{thebibliography}{4}
\expandafter\ifx\csname natexlab\endcsname\relax\def\natexlab#1{#1}\fi
\providecommand{\url}[1]{\texttt{#1}}
\providecommand{\href}[2]{#2}
\providecommand{\path}[1]{#1}
\providecommand{\DOIprefix}{doi:}
\providecommand{\ArXivprefix}{arXiv:}
\providecommand{\URLprefix}{URL: }
\providecommand{\Pubmedprefix}{pmid:}
\providecommand{\doi}[1]{\href{http://dx.doi.org/#1}{\path{#1}}}
\providecommand{\Pubmed}[1]{\href{pmid:#1}{\path{#1}}}
\providecommand{\bibinfo}[2]{#2}
\ifx\xfnm\relax \def\xfnm[#1]{\unskip,\space#1}\fi
\bibitem[{Alsofiani(2024)}]{alsofiani2024digitalization}
\bibinfo{author}{Alsofiani, M.A.}, \bibinfo{year}{2024}.
\newblock \bibinfo{title}{Digitalization in infrastructure construction projects: A prisma-based review of benefits and obstacles}.
\newblock \bibinfo{journal}{arXiv preprint arXiv:2405.16875} .
\bibitem[{Damanpour(1991)}]{damanpour1991organizational}
\bibinfo{author}{Damanpour, F.}, \bibinfo{year}{1991}.
\newblock \bibinfo{title}{Organizational innovation: A meta-analysis of effects of determinants and moderators}.
\newblock \bibinfo{journal}{Academy of management journal} \bibinfo{volume}{34}, \bibinfo{pages}{555--590}.
\bibitem[{Dibra(2015)}]{dibra2015rogers}
\bibinfo{author}{Dibra, M.}, \bibinfo{year}{2015}.
\newblock \bibinfo{title}{Rogers theory on diffusion of innovation-the most appropriate theoretical model in the study of factors influencing the integration of sustainability in tourism businesses}.
\newblock \bibinfo{journal}{Procedia-Social and Behavioral Sciences} \bibinfo{volume}{195}, \bibinfo{pages}{1453--1462}.
\bibitem[{Rogers et~al.(2014)Rogers, Singhal and Quinlan}]{rogers2014diffusion}
\bibinfo{author}{Rogers, E.M.}, \bibinfo{author}{Singhal, A.}, \bibinfo{author}{Quinlan, M.M.}, \bibinfo{year}{2014}.
\newblock \bibinfo{title}{Diffusion of innovations}, in: \bibinfo{booktitle}{An integrated approach to communication theory and research}. \bibinfo{publisher}{Routledge}, pp. \bibinfo{pages}{432--448}.

\end{thebibliography}
\end{document}